\documentclass[runningheads]{llncs}

\usepackage{eccv}

\usepackage{eccvabbrv}

\usepackage{graphicx}
\usepackage{booktabs}
\usepackage{amsmath}
\usepackage{bm}
\usepackage{caption}
\usepackage{listings}
\usepackage{multirow}
\usepackage{xcolor,colortbl}

\usepackage[accsupp]{axessibility}  

\usepackage{hyperref}

\usepackage{orcidlink}

\begin{document}

\title{3iGS: Factorised Tensorial Illumination for 3D Gaussian Splatting} 


\author{Zhe Jun Tang\inst{1}\orcidlink{0009-0002-6630-7008} \and
Tat-Jen Cham\inst{2} \orcidlink{0000-0001-5264-2572} }

\authorrunning{ZJ. Tang, TJ. Cham}

\institute{S-Lab, Nanyang Technological University \and
College of Computing \& Data Science, Nanyang Technological University \\
\email{\{zhejun001\} at \{e.ntu.edu.sg\}}\\
}

\maketitle

\begin{abstract}

The use of 3D Gaussians as representation of radiance fields has enabled high quality novel view synthesis at real-time rendering speed. However, the choice of optimising the outgoing radiance of each Gaussian independently as spherical harmonics results in unsatisfactory view dependent effects. In response to these limitations, our work, Factorised Tensorial Illumination for 3D Gaussian Splatting, or \emph{3iGS}, improves upon 3D Gaussian Splatting (3DGS) rendering quality. Instead of optimising a single outgoing radiance parameter, 3iGS enhances 3DGS view-dependent effects by expressing the outgoing radiance as a function of a local illumination field and Bidirectional Reflectance Distribution Function (BRDF) features. We optimise a continuous incident illumination field through a Tensorial Factorisation representation, while separately fine-tuning the BRDF features of each 3D Gaussian relative to this illumination field. Our methodology significantly enhances the rendering quality of specular view-dependent effects of 3DGS, while maintaining rapid training and rendering speeds.
 
  \keywords{Gaussian Splatting \and Neural Radiance Field \and Novel View Synthesis}
\end{abstract}

\section{Introduction}
\label{sec:intro}

3D Gaussian Splatting (3DGS) has emerged as the standard method for representing 3D objects and scenes, trained from images, to render photorealistic novel views. Unlike the other popular method of Neural Radiance Field (NeRF) \cite{mildenhall2020nerf}, which models a scene as an implicit continuous function, 3DGS represents surfaces with independent 3D Gaussians of different opacities, anisotropic covariances, and spherical harmonic coefficients. To render a pixel's colour, a fast, tile-based rasteriser performs alpha blending of anisotropic Gaussian splats, sorted in accordance with the visibility order.

\begin{figure*}[t]
\centering
\includegraphics[width=1.0\linewidth]{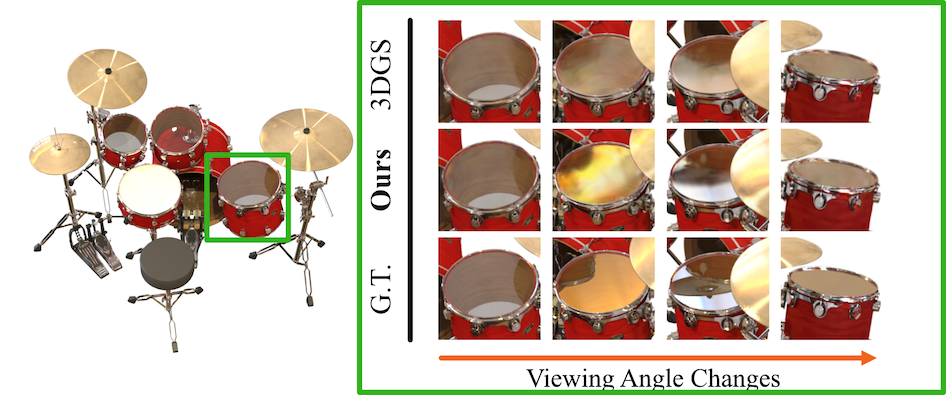}
\caption{We present test renderings from the ``Drums'' scene within the blender dataset \cite{mildenhall2020nerf}, comparing our technique against Gaussian Splatting (3DGS) \cite{kerbl20233d} and the ground truth (G.T). As the perspective shifts around the scene, the colour of the Floor Tom's top changes from translucent to reflective, showcasing intricate effects that depend on the viewpoint. These effects result from the specular reflection of incoming light and the reflections within the scene from elements like the Cymbals. Contrary to 3DGS, which struggles to capture these complex variations in light reflection, our method, 3iGS, aligns more accurately with the ground truth. }
\label{fig:teaser}
\end{figure*}

Although 3DGS shows promising performance in synthesising novel views of a scene at real-time rendering speeds, its renderings fall short in more challenging scenarios that involve complex, view-dependent surface effects. When observing images with reflective and specular surfaces, the changes in surface colour across viewing angles remain consistent, rather than exhibiting the complex variations in reflections observed in the dataset shown in Fig.\ \ref{fig:teaser}. A logical solution is to adopt the strategy of Physically Based Rendering (PBR), which involves explicitly modeling the surface characteristics and performing ray marching from surfaces to calculate illumination effects. As part of the process, the Bidirectional Reflectance Distribution Function (BRDF) of surfaces are predicted and a shading function is applied to simulate view-dependent effects \cite{cook1982reflectance, burley2012physically}. Nonetheless, accurately determining these physical properties is an ill-posed challenge, making it difficult to infer and model all the intricate rendering effects correctly.

In this paper, we draw inspiration from graphics engines that utilise illumination volumes or light probes that summarise illumination information directed towards a surface. These methods compute illumination either directly from the local illumination volume surrounding the surface \cite{greger1998irradiance} or from the nearest light probes \cite{lightprobes}, rather then sampling numerous outward rays from the surface's upper hemisphere. Such approaches allow fast rendering speed at run time, as illumination information is pre-calculated and stored in the volumes or light probes.

Our work, named Factorised Tensorial Illumination for Gaussian Splatting (3iGS), enhances 3DGS rendering quality. We introduce a continuous local illumination field of 3D Gaussians represented by compact factorised tensors for fast evaluation. The means of the 3D Gaussians serve as the input to these factorised tensors to calculate illumination features. Subsequently, each 3D Gaussian is refined through an optimisation of its mean, opacity, anisotropic covariance, diffused colour, and BRDF features. A neural renderer then maps the incident illumination neural field,  Gaussian BRDF attributes, and viewing angle to the Gaussian's specular colour. Overall, our approach represents a Gaussian's outgoing radiance as \emph{a function of both a continuous local illumination field and the individual Gaussian's BRDF attributes relative to it}. This is opposed to the conventional optimisation of the 3D Gaussians' outgoing radiance in isolation, without accounting for the effects of adjacent Gaussians or scene lighting conditions.

3iGS significantly enhances the accuracy of 3DGS, offering clear advantages in scenes with reflective surfaces where surface colours change dramatically across viewing angles as shown in Fig.\ \ref{fig:teaser}. In synthetic datasets, such as the NeRF Blender dataset and the Shiny Blender dataset, 3iGS surpasses 3DGS both quantitatively and qualitatively. Similarly, 3iGS demonstrates superior performance over 3DGS in real-world scenarios on the Tanks and Temples dataset. In summary our technical contributions are:
\begin{enumerate}
\item a method to optimise the outgoing radiance as an incident continuous illumination field and Gaussian BRDF features with a neural renderer;

\item an approach to model a continuous illumination field with Tensorial Factorisation for compactness and fast evaluation; and

\item superior performance in rendering quality over baseline 3D Gaussian Splatting while maintaining real time performance.
\end{enumerate}

\section{Related Work}
\label{sec:Related}
Our work falls into the category of learning scene representation from multi-view input images. Here we review prior work on NeRF-based representations and Gaussian splatting. We also discuss other relevant topics pertaining to inverse rendering which aims to recover scene geometry, material properties, and scene lighting conditions in Sec.\ \ref{sec:Preliminaries}.

\textbf{Scene Representations for View Synthesis} - One of the pioneering neural rendering techniques called Neural Radiance Fields (NeRF) \cite{mildenhall2020nerf} has achieved remarkable results in novel view synthesis from multi-view images. By sampling points along rays traced from the camera into the scene, NeRF reconstructs a scene as a continuous field of outgoing radiance. The technique employs volumetric rendering to determine the colour of each pixel. This method has inspired numerous developments of other scene representations \cite{mildenhall2020nerf, zhang2020nerf++, barron2021mip, barron2022mip, wang2021neus}. However, the vanilla NeRF, which encodes the entire scene representation into a set of MLPs, requires multiple queries of points along rays during training and inference. This massively slows down the speed required for real time rendering. To address this, other neural scene representation techniques apply hash encoding \cite{muller2022instant,li2023neuralangelo}, triplanes or factorised tensors \cite{chen2022tensorf, jin2023tensoir}, and gridding \cite{fridovich2022plenoxels, sun2022direct, barron2023zip} to accelerate training and inference speeds.

\textbf{Tensorial Factorisation} - In TensoRF \cite{chen2022tensorf}, a feature grid can be represented as a 4D tensor of which the first 3 represents the XYZ spatial grid and the last represents the feature channel dimension. To model a radiance field with grid representation, \cite{chen2022tensorf} propose an extension of CANDECOMP/PARAFAC (CP)-Decomposition \cite{carroll1970analysis} to Vector-Matrix (VM) decomposition:

\begin{equation}
    \label{eqn:tensorf-base}
    \begin{aligned}
    \mathcal{G}_c &= \sum_{r=1}^{R_c}\mathbf{v^X_{c,r}\circ M^{YZ}_{c,r} \circ b_{3r-2}} + \mathbf{v^Y_{c,r}\circ M^{XZ}_{c,r} \circ b_{3r-1 }} + \mathbf{v^Z_{c,r}\circ M^{XY}_{c,r} \circ b_{3r}} \\
    &= \sum_{r=1}^{R_c}\mathbf{A}_{C,r}^X \circ \mathbf{b}_{3r-2} + \mathbf{A}_{C,r}^Y \circ \mathbf{b}_{3r-1} + \mathbf{A}_{C,r}^Z \circ \mathbf{b}_{3r}
    \end{aligned}
\end{equation}

In Eq. (\ref{eqn:tensorf-base}), the inputs $\mathbf{v}$ and $\mathbf{M}$ corresponds to XYZ-mode vector and matrix factorisation and $\mathbf{b}$ denotes the appearance feature mode vectors. Separately, $\mathcal{G}_c$ and $R_C$ refers to the outgoing radiance and the colour feature channels.

\textbf{Gaussian Splatting} - As opposed to ray marching, 3D Gaussian Splatting is a recent method for rendering scenes via rasterisation. To begin, Gaussians are fitted on a point cloud that are either initialised as a set of random points or bootstrapped with a sparse point cloud produced during the SfM process for free \cite{kerbl20233d}. The Gaussians of the point cloud are defined by a function:
\begin{equation}
    \label{eqn:gs-func}
    g(\mathbf{x|\mu, \Sigma}) = e^{-\frac{1}{2}(\mathbf{x - \mu})^T\mathbf{\Sigma}^{-1}(\mathbf{x - \mu})}
\end{equation}
where each point $\mathbf{x}$ is centered at mean $\mathbf{\mu} \in \mathbb{R}^{3} $ with an anisotropic covariance matrix $\mathbf{\Sigma} \in \mathbb{R}^{3x3}$. The mean of a Gaussian is parameterised by the coordinates $\mathbf{\mu} = (\mu_x, \mu_y, \mu_z)$ that is scaled by the full 3D covariance matrix $\mathbf{\Sigma}$. As discussed in \cite{kerbl20233d}, these Gaussians have no physical meanings, given the difficulty of constraining $\mathbf{\Sigma}$ to a valid semi-positive definite matrix during the optimisation process. Instead, to derive $\mathbf{\Sigma}$, a scaling matrix $S$ and a rotation matrix $R$ is learned during the optimisation process to scale the Gaussians:
\begin{equation}
\mathbf{\Sigma} = \mathbf{RSS^TR^T} 
\end{equation}
With a viewing transformation $\mathbf{W}$ and an affine approximation of the projective transformation $\mathbf{J}$, the covariance matrix is then expressed in camera coordinates as:
\begin{equation}
\mathbf{\Sigma^{'} = JW\Sigma W^TJ^T}
\end{equation}
Furthermore, each Gaussian is coloured via a set of Spherical Harmonics (SH) coefficients that represent the view dependent colour $c_i$, also known as radiance field, multiplied by its opacity $\alpha$. To colour a pixel $u$ as $\hat{C}$, alpha blending of $N$ ordered Gaussians is applied:
\begin{equation}
\hat{\mathbf{C}} = \sum_{i\in N}T_i g_i(\mathbf{u | \mu^{'},\mathbf{\Sigma^{'}}})\alpha_i\mathbf{c}_i ,  \ \ T_i=\prod^{i-1}_{j=1}(1-{g}_i(\mathbf{u}|\mathbf{\mu}', \mathbf{\Sigma}')\alpha_i)
\end{equation}

\subsection{Preliminaries}\label{sec:Preliminaries}

As discussed, the direct optimisation of spherical harmonics to describe the outgoing radiance in individual Gaussians in 3DGS results in poor view-dependent effects. A crucial reason is that these Gaussians do not fully model scene properties \cite{jiang2023gaussianshader} and thus fail to capture the specular effects which changes drastically across viewing angles. 

Therefore to account for the specular highlights, it is beneficial to model the underlying  properties such as the BRDF and illumination effects of the scene. In conventional computer graphics, a rendering equation is commonly applied to simulate effects of specular and diffused shading \cite{kajiya1986rendering}. For instance, rendering Eq.\ (\ref{eq:rendering-eq}) describes an outgoing radiance of a surface point:
\begin{equation}
    \label{eq:rendering-eq}
    L_o(\mathbf{x}, \mathbf{v}) = \int_\Omega L_i(\mathbf{x}, \mathbf{l}) f_r(\mathbf{l}, \mathbf{v}) (\mathbf{l} \cdot \mathbf{n}) d\mathbf{l},
\end{equation}
The radiance $L_o$ emitted from a surface point $\mathbf{x}$, when observed from a viewing direction $\mathbf{v}$, is defined in Eq.\ (\ref{eq:rendering-eq}). An integral is applied to accumulate the contribution of incident light at an incident angle $\mathbf{l}$ across the upper hemisphere $\mathbf{\Omega}$ of $\mathbf{x}$. The function $f_r$ denotes the Bidirectional Radiance Distribution Function (BRDF), describing the reflection characteristics of incident radiance at $\mathbf{x}$ viewed in direction $\mathbf{v}$. Lastly the inclusion of the cosine law with the normal vector $\mathbf{n}$ ensures the energy conservation.

From a signal processing perspective, an alternative to Eq.\ (\ref{eq:rendering-eq}) is expressed more generally in terms of spherical harmonic convolution \cite{mahajan2007theory, kautz2002fast}:
\begin{equation}
    \label{eq:sh-rendering}
    B_{lm} = \Lambda_{l}\rho_lL_{lm}
\end{equation}
In Eq.\ (\ref{eq:sh-rendering}), $B_{lm}$ defines the outgoing reflected light as the product of BRDF filter $\rho_l$, spherical harmonic coefficients of lighting signal $L_{lm}$, and the normalisation constant $\Lambda_{l}$.

Some studies \cite{jiang2023gaussianshader, shi2023gir} enhance 3DGS by expressing BRDF $f_r$ as a Cook-Torrance microfacet model \cite{cook1982reflectance} or the GGX Trowbridge-Reitz model \cite{burley2012physically}. In these approaches, physical attributes, including roughness $r$, albedo $a$, metallicity $m$, and the normal vector $\mathbf{n}$ are predicted and used in Eq.\ (\ref{eq:rendering-eq}). Although these modifications marginally improve rendering quality metrics, they fail to accurately produce high-quality, view-dependent effects. This shortfall primarily stems from relying on estimated parameters for physical rendering within a simplified rendering equation \cite{liang2023envidr}. Furthermore, these parameters are inherently challenging to be estimated accurately, due to the ill-posed nature of inverse rendering from multi-view images. Although numerous works \cite{boss2021neural, liang2023envidr,jin2023tensoir,boss2021nerd, bi2020neural, srinivasan2021nerv} also achieved success by exploring a neural representation of the rendering equation, these works either require prior information, such as known lighting conditions or a pre-trained model on a realistic dataset with known BRDF parameters. Furthermore these techniques are experimented with ray tracing based methods like NeRF. A work closest to ours in the area of rasterisation and Gaussian Splatting manner, is GaussianShader \cite{jiang2023gaussianshader} which we compare against in Sec.\ \ref{sec:comparisons}. 

\section{Method}
\label{sec:method}

Instead of predicting the physical BRDF properties of materials in the scene, our goal is to express the outgoing radiance of a Gaussian as a more general expression of BRDF, and the incoming illumination as neural features. This idea is based on a generalized version of Eq.\ (\ref{eq:sh-rendering}), where BRDF features modify an incoming illumination field, without the need for decomposing down to intrinsic material properties \cite{mahajan2007theory}. 

\begin{figure*}[t]
\centering
\includegraphics[width=1.0\linewidth]{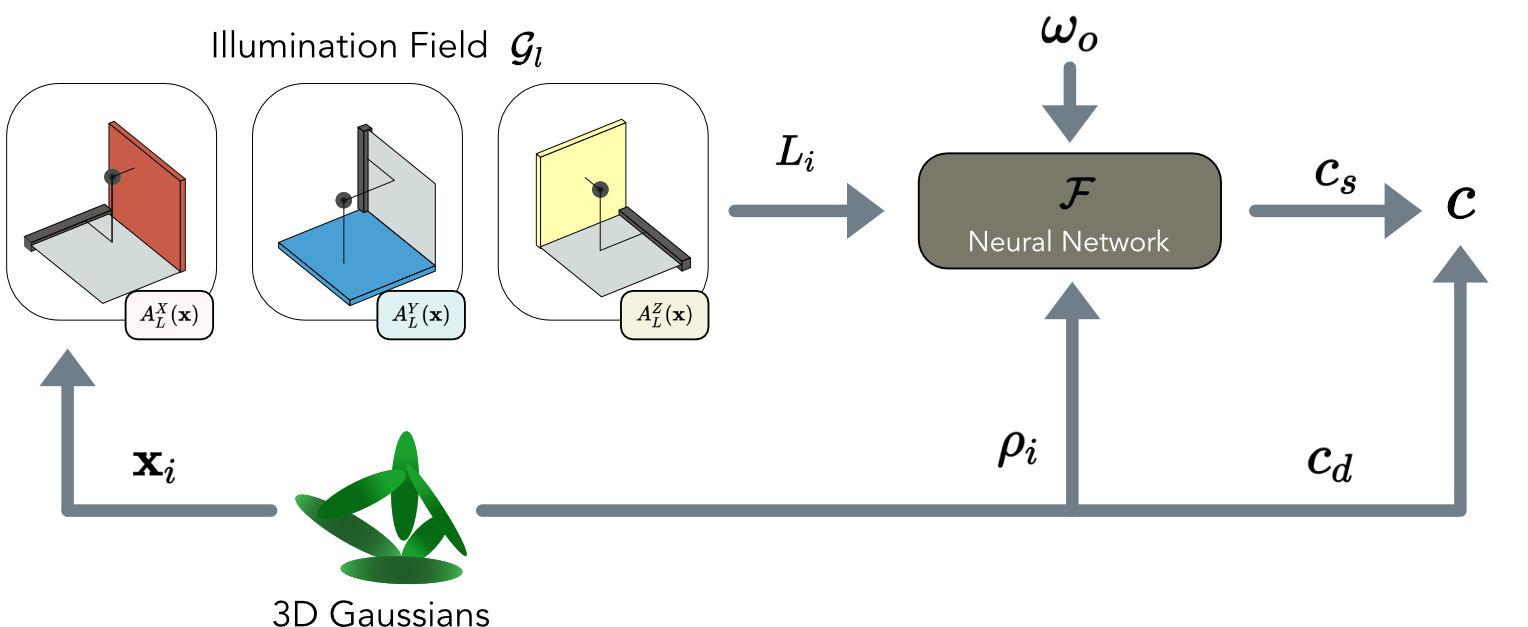}
\caption{A visualisation of 3iGS pipeline to render a single Gaussian's colour. We interpolate an incident illumination $L_{i}$ from the factorised tensorial illumination field $\mathcal{G}_l$ using a Gaussian mean $\bm{x}_i$ as input. A neural network $\mathcal{F}$ maps the illumination field $L_{i}$, the Gaussian BRDF features $\rho_i$, and the viewing direction $\omega_o$ to Gaussian's specular colour $c_s$. Following, the diffused colour $c_d$ and specular colour $c_s$ are added linearly to produce the final outgoing radiance field $c$.}
\label{fig:neuralengine}
\end{figure*}

Specifically for each 3D Gaussian in the scene, the outgoing radiance field is formed by:
\begin{equation}
    \label{eq:shade-gaussian}
   \mathbf{c}(\omega_o) = \mathbf{c_d} + \mathbf{c_s}(\omega_o)
\end{equation}
For viewing angle $\omega_o$, a Gaussian is coloured by its constant diffused colour $c_d$ and a view dependent specular colour ${c_s}$. At each Gaussian $i$, a small neural network $\mathcal{F}$ maps the Gaussian BRDF features $\rho_i$ and the incoming illumination $ L_{i}$ to its specular colour viewed at an angle  $\omega_o$ :
\begin{equation}
    \label{eq:spec_eqn}
    \mathcal{F} : \{ \rho_i, L_{i}, \omega_o \} \mapsto {\mathbf{c_s}}
\end{equation}

\subsection{Illumination Grid by Tensorial Factorisation}
\label{sec:illu-grid}
Our work is largely inspired by conventional computer graphics engines for fast rendering of scene and objects in video games. The fundamental rendering equation highlights the role of multi-bounce lighting in achieving indirect illumination, wherein light bounces off one surface to illuminate another. However, the process of ray tracing from each Gaussian surface into the scene is notably resource-intensive, undermining the goal of quick rendering in 3D graphics systems. To facilitate real-time rendering, one strategy involves the use of baking techniques that employ irradiance volumes \cite{greger1998irradiance}. This method segments a scene into distinct volumes and pre-calculates irradiance data offline. An alternative strategy places light probes \cite{lightprobes, reflectionprobes} throughout the scene to gather lighting information at specific spatial locations. When rendering the colour of a surface, the system quickly interpolates lighting information from the nearest light probes, ensuring swift rendering times.

To maintain the fast rendering speed of 3DGS, our work describes a methodology of learning the illumination features of a Gaussian with a continuous grid based illumination field as:
\begin{equation}
    \label{eq:illumination_triplane}
    \mathcal{G}_l : \{\mathbf{x_i} \} \mapsto L_{i}
\end{equation}
Given a Gaussian's mean coordinate $\mathbf{x_i}$, we seek to compute an illumination field $L_{i}$ by interpolating from learnable grid representation. The illumination tensors $\mathcal{G}_l$ is formulated similar to TensoRF  \cite{chen2022tensorf} by a vector-matrix spatial factorisation as follows:
\begin{equation}
    \label{eqn:tensorf}
    \mathcal{G}_l = \sum_{r=1}^{R_L}\mathbf{A}_{L,r}^X \circ \mathbf{b}_{3r-2} + \mathbf{A}_{L,r}^Y \circ \mathbf{b}_{3r-1} + \mathbf{A}_{L,r}^Z \circ \mathbf{b}_{3r}
\end{equation}
In Eq.\ (\ref{eqn:tensorf}), $R_L$ represents the feature channels of the illumination components, $\bm{A}$ as feature tensors and $\mathbf{b}$ as feature vectors. The illumination feature grid is jointly learned end to end in the optimisation process together with each Gaussian in the scene. Unlike 3DGS, where each Gaussian is optimised independently, the illumination field is modelled as a continuous grid function. A Gaussian mean serves as the input to query from the factorised tensor grid via interpolation. The inclusion of this continuous incoming illumination field directed at each Gaussian is the core component of producing accurate view-dependent effects, as we show in the ablation study of Sec.\ \ref{sec:ablation}. Furthermore, by formulating this field as factorised tensors, it allows the network to achieve fast rendering speed. Our illumination field is coarse, using 87.5\% less voxels compared to TensoRF on synthetic datasets. This compact representation is also low in memory footprint compared to the number of optimised Gaussians, which is often a magnitude order or more higher. We refer readers to \cite{chen2022tensorf}, which provides a comprehensive overview to describe how the tensors are factorised and interpolated.

\subsection{3D Gaussian Features}
\label{sec:gaussian-features}
In 3DGS \cite{kerbl20233d}, Gaussians are optimised with a set of parameters: 3D positions, opacity $\alpha$, anisotropic covariance, and spherical harmonics coefficients. In our work, instead of optimising spherical harmonics as an outgoing radiance, 3iGS characterises the Gaussians with a diffused colour and learnable BRDF features. Unlike \cite{jiang2023gaussianshader, shi2023gir}, we do not strictly enforce physically interpretable properties commonly used in shading techniques. Aforementioned, these techniques are often simplified, too ill-posed to be decomposed individually, and insufficient to encompass all complex rendering effects \cite{liang2023envidr}. Rather, we loosely follow Eq.\ (\ref{eq:sh-rendering}) and treat BRDF feature components as a set of weights that alter the incoming illumination field. Given a continuous illumination field obtained from Eq.\ (\ref{eqn:tensorf}), a Gaussian's BRDF is conditionally optimised against it. This is in contrast to 3DGS where the Gaussians' outgoing radiance are individually optimised without modelling the interdependencies that should arise from a shared scene illumination, resulting in detrimental view-dependent effects.

\subsection{Shading Gaussians}
\label{sec:shading-gaussians}

Following Eq.\ (\ref{eq:spec_eqn}), we shade each Gaussian by mapping its viewing directions encoded with Integrated Directional Encoding (IDE) \cite{verbin2022ref}, Gaussian features (obtained in Sec.\ \ref{sec:gaussian-features}), and its illumination field, to the specular colour output. We linearly add the diffused and specular colours to create its radiance field as per Eq.\ (\ref{eq:shade-gaussian}). To render the final scene, we follow the rasterisation pipeline proposed in the original 3DGS work.

\section{Optimisation}
\label{sec:optimisation}
In the previous Sec.\ \ref{sec:method}, we described the necessary components to model a scene with Gaussians and render it via rasterisation. To improve the stability of training and to enhance the final rendering quality, we first train the model with the diffused colour in the first 3,000 iterations. Following, specular colours are added to the Gaussians as in Eq.\ (\ref{eq:shade-gaussian}). 

While training the tensorial illumination grid, an initial boundary which encapsulate the scene bounding box is defined. Midway through training, we shrink the illumination grid to fit the Gaussians and resample the grid with the same number of voxels. We adopt the same adaptive control of Gaussians of 3DGS\cite{kerbl20233d} to limit the number of Gaussians and the units per volume. We propose to train our model with the same loss function as 3DGS for a fair evaluation: 
\begin{equation}
    \label{eqn:loss}
    \mathcal{L} = (1 - \lambda )\mathcal{L}_1 + \lambda\mathcal{L}_{\text{D-SSIM}}
\end{equation}
where we combined the $\mathcal{L}_1$ term with a D-SSIM term with $\lambda$ set to 0.2.

\section{Experiments and Results}

\subsection{Datasets}
\textbf{Synthetic scenes} - We show experimental results of 3iGS based on the Blender dataset released in \cite{mildenhall2020nerf}. This dataset contains challenging scenes of complex geometries with realistic non-Lambertian materials. Similarly, we evaluate our model on the Shiny Blender dataset presented in \cite{verbin2022ref}. Unlike the Blender dataset, Shiny Blender contains a singular object with simple geometries in each scene with more glossy effects.\\
\textbf{Real world complex scenes} - To prove the effectiveness of our model in real world scenes, we evaluate our renderings on the Tanks and Temples dataset \cite{Knapitsch2017}. This dataset is obtained from video sequences of real world objects and environment.

\subsection{Comparisons} \label{sec:comparisons}

To evaluate our model, we compared against methods that apply both ray-tracing methods like NeRF, or rasterisation methods with Gaussian Splatting. Out of all prior work, 3DGS and GaussianShader is the closest work which offers real time inference speed which we will mainly compare against. On comparing the qualitative result figures, we re-ran the experiments of 3DGS \cite{kerbl20233d} and GaussianShader \cite{jiang2023gaussianshader} using their original repository code under settings specified by the authors. \textbf{Ray-Tracing Methods} such as \cite{mildenhall2020nerf, verbin2022ref, liang2023envidr} represent a scene as a radiance field using MLPs. By performing multiple samplings on rays marched from the camera into the scene, the sampled points are queried with MLP to obtain the opacity and radiance values. Volume rendering is performed to obtain the final pixel colour. \textbf{Rasterisation Methods} such as Gaussian Splatting (3DGS) \cite{kerbl20233d} and GaussianShader \cite{jiang2023gaussianshader} apply a rasterisation pipeline as opposed to ray tracing methods. These models represents a scene as Gaussians with radiance properties based on Spherical Harmonics. In, \cite{jiang2023gaussianshader}, 3DGS is extended by modelling a scene with additional material characteristics and a shading function is applied, as opposed to ours which uses an MLP as neural renderer. Furthermore, \cite{jiang2023gaussianshader} shades Gaussians with a global differentiable environment light stored in cube maps, and optimises independent Gaussians with spherical harmonic-based color for unaccounted illumination. In our work, we represent incident illumination \emph{locally} with grid-based tensors and optimise Gaussian BRDF features relative to this field.

For a fair comparison, 3iGS is trained with the same loss function as 3DGS as described in Sec.\ \ref{sec:optimisation} and the same number of iterations of 30,000 steps. We repurposed the 16$\times$3 SH coefficients in 3DGS as BRDF feature channels and added 4 additional parameters of base colour and roughness for IDE view-directional encoding. The tensorial illumination field is set at a coarse resolution size of $150^3$ voxels.

\begin{figure}[h]
\centering
\includegraphics[width=0.8\linewidth]{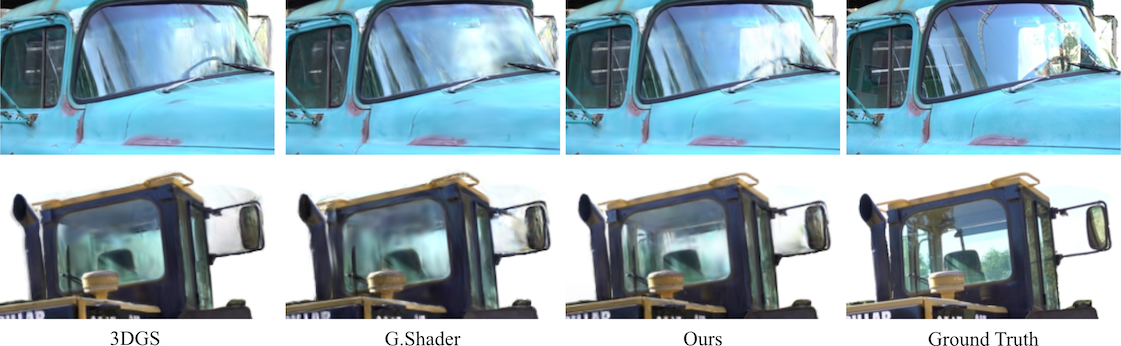}
\caption{Comparisons of test-set views of real world scenes. 3iGS enhances 3DGS renderings by producing clearer view dependent effects as shown.}
\label{fig:tanks-temples}
\end{figure}


\begin{table}[htbp]
    \caption{Our approach demonstrates superior quantitative performance over current methods when tested on Synthetic Datasets. Specifically, within the NeRF Synthetic dataset, our method surpasses all competitors across various image quality assessments (PSNR/SSIM/LPIPS). In the context of the Shiny Blender dataset, 3iGS matches the performance of existing rasterization techniques in terms of PSNR and SSIM but surpasses them in LPIPS for the majority of scenes. We encourage readers to examine the accompanying figure showcasing renderings of the Shiny Blender scene, where our method attains enhanced qualitative outcomes. Best results, benchmarked across \emph{real time rendering methods}, are in bold.}
        \centering
        \small

    \resizebox{\columnwidth}{!}{ 
    \begin{tabular}{l*{9}{p{1.4cm}}}
    \hline
    \multicolumn{10}{c}{NeRF Synthetic \cite{mildenhall2020nerf}}\\
    \hline
     & Chair & Drums & Lego & Mic & Mats. & Ship & Hotdog & Ficus & Avg.  \\ \hline
    \multicolumn{10}{c}{PSNR$\uparrow$}\\ \hline
    
    NeRF~\cite{mildenhall2020nerf} & 33.00 & 25.01 & 32.54 & 32.91 & 29.62 & 28.65 & 36.18 & 30.13 & 31.01 \\
    Ref-NeRF~\cite{verbin2022ref} & 33.98 & 25.43 & 35.10 & 33.65 & 27.10 & 29.24 & 37.04 & 28.74 & 31.29 \\
    ENVIDR~\cite{liang2023envidr} & 31.22 & 22.99 & 29.55 & 32.17 & 29.52 & 21.57 & 31.44 & 26.60 & 28.13 \\ \hline
    3DGS~\cite{kerbl20233d} & 35.82 & 26.17 & 35.69 & 35.34 & 30.00 & 30.87 & 37.67 & 34.83 & 33.30 \\
    G.Shader~\cite{jiang2023gaussianshader} & 35.83 & 26.36 & 35.87 & 35.23 & \textbf{30.07} & 30.82 & 37.85 & 34.97 & 33.38 \\ 

    G.Shader(reproduced)~\cite{jiang2023gaussianshader} & 33.70 & 25.50 & 32.99 & 34.07 & 28.87 & 28.37 & 35.29 & 33.05 & 31.48 \\ 
    
    Ours &  \textbf{35.90} & \textbf{26.75} & \textbf{35.94} & \textbf{36.01} & 30.00 & \textbf{31.12} & \textbf{37.98} & \textbf{35.40} & \textbf{33.64} \\
    
    \hline \hline
    \multicolumn{10}{c}{SSIM$\uparrow$}\\ \hline
    
    NeRF~\cite{mildenhall2020nerf} & 0.967 & 0.925 & 0.961 & 0.980 & 0.949 & 0.856 & 0.974 & 0.964 & 0.947 \\
    Ref-NeRF~\cite{verbin2022ref} & 0.974 & 0.929 & 0.975 & 0.983 & 0.921 & 0.864 & 0.979 & 0.954 & 0.947 \\
    ENVIDR~\cite{liang2023envidr} & 0.976 & 0.930 & 0.961 & 0.984& 0.968 & 0.855 & 0.963 & 0.987 & 0.956 \\ \hline
    3DGS~\cite{kerbl20233d} & 0.987 & 0.954 & 0.983 & 0.991 & 0.960 & 0.907 & 0.985 & 0.987 & 0.969 \\
    G.Shader~\cite{jiang2023gaussianshader} & 0.987 & 0.949 & 0.983 & 0.991 & 0.960 & 0.905 & 0.985 & 0.985 & 0.968 \\ 
    
    G.Shader(reproduced)~\cite{jiang2023gaussianshader} & 0.980 & 0.945 & 0.972 & 0.989 & 0.951 & 0.881 & 0.980 & 0.982 & 0.960\\ 
    
    Ours & \textbf{0.987} & \textbf{0.955} & \textbf{0.983} & \textbf{0.992} & \textbf{0.961} & \textbf{0.908} & \textbf{0.986} & \textbf{0.989} & \textbf{0.970} \\
    \hline \hline
    \multicolumn{10}{c}{LPIPS$\downarrow$}\\ \hline
    
    NeRF~\cite{mildenhall2020nerf} & 0.046 & 0.091 & 0.050 & 0.028 & 0.063 & 0.206 & 0.121 & 0.044 & 0.081 \\
    Ref-NeRF~\cite{verbin2022ref} & 0.029 & 0.073 & 0.025 & 0.018 & 0.078 & 0.158 & 0.028 & 0.056 & 0.058 \\
    ENVIDR~\cite{liang2023envidr} & 0.031 & 0.080 &0.054 &0.021 & 0.045 & 0.228 & 0.072 & 0.010 & 0.067 \\ \hline
    3DGS~\cite{kerbl20233d} & 0.012 & 0.037 & 0.016 & 0.006 & 0.034 & 0.106 & 0.020 & 0.012 & 0.030 \\
    
    G.Shader~\cite{jiang2023gaussianshader} & 0.012 & 0.040 & \textbf{0.014} & 0.006 &\textbf{0.033} &\textbf{0.098} & 0.019 & 0.013 & 0.029 \\

    G.Shader(reproduced)~\cite{jiang2023gaussianshader} & 0.019 & 0.045 & 0.026 & 0.009 & 0.046 & 0.148 & 0.029 & 0.017 & 0.042\\
    
    Ours &\textbf{0.012} & \textbf{0.036} & 0.015 & \textbf{0.005} & 0.034 & 0.102 &\textbf{0.019} & \textbf{0.010} & \textbf{0.029} \\
    \hline 
    \end{tabular}
    }

    \resizebox{\columnwidth}{!}{
    \begin{tabular}{l*{8}{p{1.7cm}}}
    \hline
    \multicolumn{8}{c}{Shiny Blender~\cite{verbin2022ref}} \\
             & Car & Ball & Helmet & Teapot & Toaster & Coffee & Avg.\\ \hline
             
    \multicolumn{8}{c}{PSNR$\uparrow$} \\ \hline
    
    NVDiffRec~\cite{munkberg2022extracting} &  27.98 & 21.77 & 26.97 & 40.44 & 24.31 & 30.74 & 28.70 \\
    Ref-NeRF~\cite{verbin2022ref} &  30.41 &  29.14 &  29.92 & 45.19 &  25.29 & 33.99 &  32.32 \\
    ENVIDR~\cite{liang2023envidr} & 28.46 & 38.89 & 32.73 & 41.59 & 26.11 & 29.48 & 32.88 \\ \hline
    3DGS~\cite{kerbl20233d} & 27.24 & 27.69 & 28.32 &  45.68 & 20.99 & 32.32 & 30.37 \\
    G.Shader~\cite{jiang2023gaussianshader} & \textbf{27.90} & \textbf{30.98} & 28.32 & 45.86 & \textbf{26.21} &  32.39 & \textbf{31.94} \\

    G.Shader(reproduced)~\cite{jiang2023gaussianshader} & 27.51 & 29.02 & \textbf{28.73} & 43.05 & 22.86 &  31.34 & 30.41 \\

    Ours & 27.51 & 27.64 & 28.21 & \textbf{46.04} & 22.69 & \textbf{32.58} & 30.77\\
    \hline 
    \hline
    
    \multicolumn{8}{c}{SSIM$\uparrow$} \\ \hline
    
    NVDiffRec~\cite{munkberg2022extracting} & 0.963 & 0.858 &  0.951 & 0.996 &  0.928 &  0.973 & 0.945 \\
    Ref-NeRF~\cite{verbin2022ref} &   0.949 &  0.956 & 0.955 & 0.995 & 0.910 & 0.972 &  0.956 \\
    ENVIDR~\cite{liang2023envidr} &  0.961 & 0.991 & 0.980 & 0.996 & 0.939 & 0.949 & 0.969 \\ \hline
    3DGS~\cite{kerbl20233d} & 0.930 & 0.937 &  0.951 & 0.996 & 0.895 & 0.971 & 0.947 \\
    G.Shader~\cite{jiang2023gaussianshader} &\textbf{0.931} & \textbf{0.965}& 0.950 & 0.996 & \textbf{0.929} & 0.971 & \textbf{0.957} \\ 

    G.Shader(reproduced)~\cite{jiang2023gaussianshader} &0.930& 0.954& \textbf{0.955} & 0.995  & 0.900 &  0.969 & 0.950 \\

    Ours & 0.930 & 0.938 &  0.951 & \textbf{0.997} & 0.908 & \textbf{0.973} & 0.949 \\
    \hline \hline
    
    \multicolumn{8}{c}{LPIPS$\downarrow$} \\ \hline
    
    NVDiffRec~\cite{munkberg2022extracting} & 0.045 & 0.297 & 0.118 & 0.011 & 0.169 & 0.076 & 0.119 \\
    Ref-NeRF~\cite{verbin2022ref} & 0.051 & 0.307 & 0.087 & 0.013 & 0.118 & 0.082 & 0.109 \\
    ENVIDR~\cite{liang2023envidr} & 0.049 &  0.067 & 0.051 & 0.011 &  0.116 & 0.139 & 0.072 \\ \hline
    3DGS~\cite{kerbl20233d} & 0.047 & 0.161 & 0.079 & 0.007 & 0.126 &  0.078 & 0.083 \\
    G.Shader~\cite{jiang2023gaussianshader} & 0.045 & \textbf{0.121} &  0.076 & 0.007 & \textbf{0.079} &  0.078 & \textbf{0.068} \\ 

    G.Shader(reproduced)~\cite{jiang2023gaussianshader} & 0.045 & 0.148 & 0.088 & 0.012 & 0.111&  0.085 & 0.099 \\

    Ours & \textbf{0.045} & 0.156 &  \textbf{0.073} & \textbf{0.006} & 0.099 & \textbf{0.076} &  0.075 \\
    \hline \hline
    \end{tabular}
    }
    
    \label{tab:test}
\end{table}

\begin{figure}[htbp]
\centering
\includegraphics[scale=0.60]{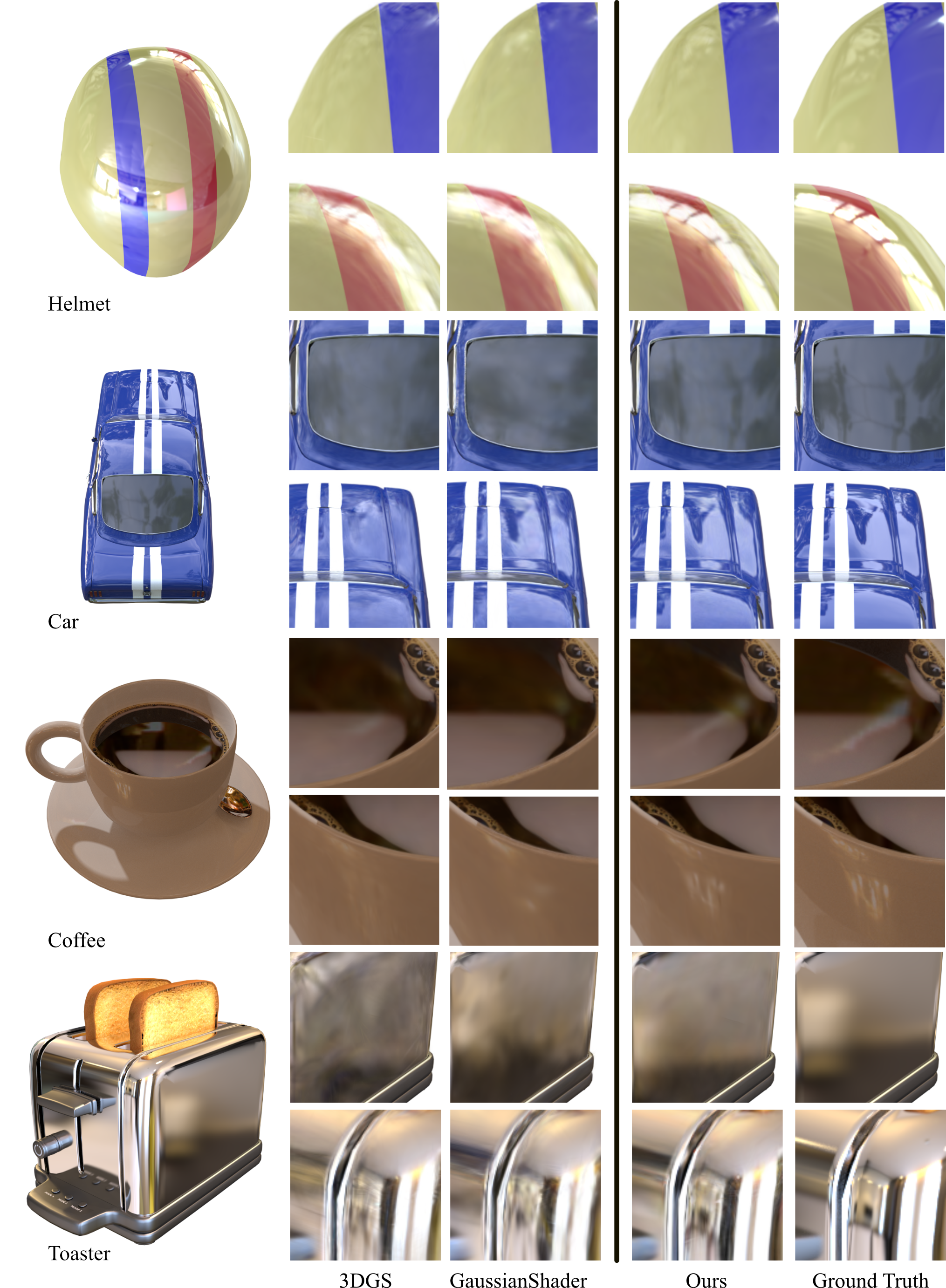}
\caption{In evaluating test-set views from the Shiny Blender dataset, we compared the performance of 3DGS \cite{kerbl20233d}, GaussianShader \cite{jiang2023gaussianshader}, and our work 3iGS. The standard 3DGS method generally yields the least satisfactory renderings, with images often appearing blurry in areas of specular reflection. GaussianShader shows a slight improvement by incorporating the GGX BRDF model, leading to marginally better results in rendering specular regions. In contrast, 3iGS stands out by employing a general rendering function that predicts neural features of illumination field and BRDF instead of relying on physical parameters. This approach allows 3iGS to surpass existing methods significantly, capturing the intricate details within specular highlights with remarkable precision.}
\label{fig:shiny_blender}
\end{figure}

\clearpage

\begin{table}[htbp]
    \caption{A quantitative comparisons (PSNR / SSIM / LPIPS) between 3DGS \cite{kerbl20233d}, GaussianShader \cite{jiang2023gaussianshader}, and our method on real world scenarios on Tanks and Temples Dataset \cite{Knapitsch2017}}.
    \centering
    \resizebox{\columnwidth}{!}{ 

    \begin{tabular}{l|*{7}{p{1.8cm}}}
    \hline
    \multicolumn{7}{c}{Tanks and Temples Dataset~\cite{Knapitsch2017}} \\
             & Barn & Caterpillar & Family & Ignatius & Truck & Avg.\\ \hline 
             
    \multicolumn{7}{c}{PSNR$\uparrow$} \\ \hline
    3DGS~\cite{kerbl20233d} & 29.13 & 26.17 & 34.88 & 29.50  &28.38 & 29.61 \\
    G.Shader(reproduced)~\cite{jiang2023gaussianshader} & 27.67 & 25.23 & 33.52 & 28.28 & 27.61 & 28.46 \\
    Ours & \textbf{29.73} &  \textbf{27.04} &  \textbf{35.36} &  \textbf{30.04} &  \textbf{28.82} &  \textbf{30.20} \\ \hline \hline 

    \multicolumn{7}{c}{SSIM$\uparrow$} \\ \hline
    3DGS~\cite{kerbl20233d} & 0.920 & 0.932 & 0.982 & 0.973 & 0.945 & 0.950 \\
    G.Shader(reproduced)~\cite{jiang2023gaussianshader} & 0.897 & 0.915 & 0.977 & 0.968 & 0.935 & 0.938 \\
    Ours &  \textbf{0.923} &  \textbf{0.938} &  \textbf{0.983} &  \textbf{0.974} &  \textbf{0.947} &  \textbf{0.953} \\ \hline \hline

    \multicolumn{7}{c}{LPIPS$\downarrow$} \\ \hline
    3DGS~\cite{kerbl20233d} & 0.113 & 0.074 & 0.023 & 0.032 & 0.059 & 0.060 \\
    G.Shader(reproduced)~\cite{jiang2023gaussianshader} & 0.147 & 0.098 & 0.029 & 0.039 & 0.071 & 0.077 \\
    Ours &  \textbf{0.112} &  \textbf{0.071} &  \textbf{0.022} &  \textbf{0.031} &  \textbf{0.057} & \textbf{0.058} \\
    \hline \hline
    \end{tabular}
}
\end{table}

\begin{figure}[htbp]
\centering
\includegraphics[width=1.0\linewidth]{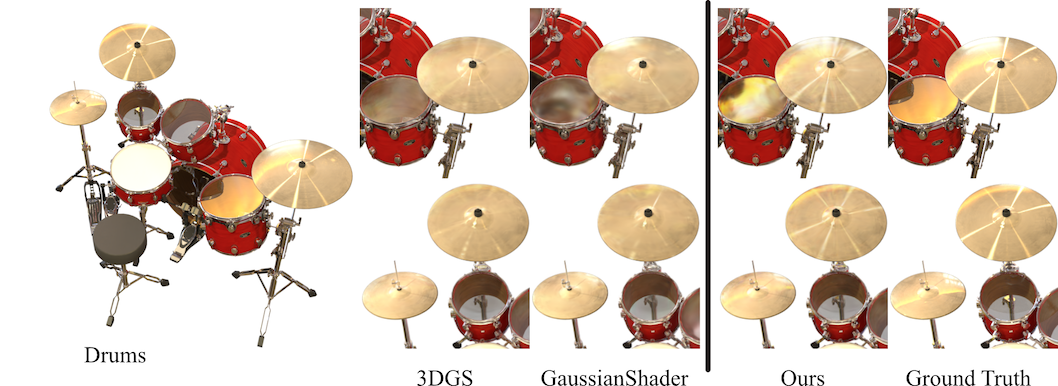}
\caption{ In contrast to 3DGS \cite{kerbl20233d} and GaussianShader \cite{jiang2023gaussianshader}, our 3iGS method uniquely identifies both the golden specular highlights and the reflections on the Medium Tom as seen in the plastic surface of the Floor Tom (top row). Our approach successfully captures the detailed specular highlights on every cymbal within the drum setup from the Blender dataset, as presented in \cite{mildenhall2020nerf}.}
\label{fig:blender}
\end{figure}

\subsection{Discussion}
In the comparisons detailed in Sec.\ \ref{sec:comparisons}, 3iGS demonstrates superior performance over the established baselines, delivering both quantitatively and qualitatively enhanced renderings in a majority of test cases on real time rendering rasterisation approaches. In the NeRF Synthetic dataset, 3iGS surpasses the prior 3DGS and GaussianShader. Although GaussianShader reportedly performs slightly better on the Shiny Blender dataset, we have included both reported and reproduced results based on the official code repository from the authors. We postulate that the Shiny Blender dataset scenes, which comprise single objects only, presents simpler geometries which facilitates an easier recovery of intrinsic material properties essential for rendering view-dependent effects. In addition, specular reflections in this dataset is primarily dominated by direct illumination from an external environment map. Thus GaussianShader which models direct lighting with a differentiable environment cube map performs well. However, when presented with a complex scene containing multiple objects, such as the NeRF Synthetic dataset shown in Fig.\ \ref{fig:teaser} with its intricate intra-scene interactions, GaussianShader struggles to accurately recover the physical rendering parameters.
Furthermore these lighting scenarios are more complex due to indirect lighting. Therefore jointly modelling direct and indirect lighting using a continuous local incident field is crucial. NeRF based approaches reported above present competitive results. Yet, such methods are extremely slow to train, often requiring days, and are unable to perform real-time rendering needed for interactive applications.

Comparing across all methodologies, our 3iGS method presents an attractive and pragmatic alternative to achieve excellent rendering quality while balancing rendering speed, as discussed in Sec.\ \ref{sec:ablation}.

\subsection{Ablation Studies} \label{sec:ablation}

\begin{table}[htbp]
    \caption{An ablation study of our model on the Blender synthetic dataset. We experiment 3iGS under a variety of model parameters. In the first row, we directly an outgoing radiance field similar to NeRF based methods. The second row omits the prediction of a BRDF roughness parameter which encodes the viewing direction as IDE. Both experimental results are inferior compared to our complete model.}
    \centering
    \small
    \begin{tabular}{ l| *{3}{c}} & PSNR & SSIM & LPIPS \\
    \hline

Ours (outgoing radiance field) & 32.38 & 0.965 & 0.035 \\
Ours (no roughness parameter, i.e IDE) & 33.26 & 0.967 & 0.031 \\
Ours (complete model) & \textbf{33.64} & \textbf{0.970} & \textbf{0.029} \\
    
    \end{tabular}%
    \label{tab:ablation_table}
\end{table}

\begin{table}[htbp]
    \centering
    \small
    \caption{We evaluate the test and train speed of 3DGS \cite{kerbl20233d} and GaussianShader \cite{jiang2023gaussianshader} on a single Tesla V100 32Gb VRAM GPU with the original codebase and settings advocated by the authors. We then report the results normalised with these rendering speed of 3DGS.}

    \begin{tabular}{ l| *{2}{c}} & Test & Train \\
    \hline
    3DGS & 1.0x & 1.0x \\
    GaussianShader & 6.3x slower & 12.1x slower \\
    Ours & 2.0x slower & 3.2x Slower \\
    \end{tabular}%
    \label{tab:speed}

\end{table}
 
In Tab.\ \ref{tab:ablation_table}, we study the effectiveness of our design choices and parameters for 3iGS. In the first row, we use the Gaussian mean and interpolate features from the factorised tensors and predict the outgoing specular colours directly. In this scenario, we predict the outgoing radiance field similar to a NeRF like manner for specular colours. In the second row, we abandon the BRDF roughness parameters from the Gaussian features and apply a standard Fourier positional encoding of viewing direction. Both cases led to inferior renderings as compared to our complete model.

In Tab.\ \ref{tab:speed}, we illustrate the training and rendering speed (test) of 3iGS against 3DGS and GaussianShader. We normalise the speed based on 3DGS. Our model performs competitively and achieve real time rendering speed although it is slower than 3DGS whereas GaussianShader performs much slower than the vanilla model. We attribute the efficient rendering speed to the use of factorised tensors for the illumination field.

\section{Limitations and Weaknesses}
3iGS inherits the main challenges of factorised tensors as \cite{chen2022tensorf}. Our model is limited to scenes that fit within a defined bounding box. Future works could explore this direction in warping unbounded scenes to fit a tensorial grid representation. Furthermore, 3iGS inherits the weaknesses of 3DGS; a large VRAM GPU is necessary to fit 3D Gaussians, and to evaluate the illumination field. A straightforward workaround is to reduce the number of Gaussians created by adding an upper bound on the number of produced Gaussians in the adaptive control step. Our work also inherits 3DGS's difficulty in producing accurate scene geometry.

\section{Conclusion}
We introduce our work, \emph{Factorised Tensorial Illumination for 3D Gaussian Splatting (3iGS)}, to enhance the view-dependent effects in rendering Gaussian radiance fields. Our approach overcomes the constraints of previous methods, which relied on optimising an outgoing radiance field of independent Gaussians with Spherical Harmonics (SH) parameters. We illustrate that superior view-dependent effects in 3DGS can be attained by depicting an outgoing radiance field as a continuous illumination field and the Gaussian's BRDF characteristics in relation to this field. Distinct from other methods depending on over-simplified yet restrictive rendering equations that require prediction of physical attributes of scene surfaces for shading, our methodology proves to be more efficacious. Furthermore, we have shown that fast rendering speeds are attainable through the representation of an illumination field with factorised tensors. We demonstrated our claims across diverse datasets, from synthetic to real-world environments, and compared against prior art on both quantitative and qualitative metrics. We also evaluate the effectiveness of our model parameters and design choices through an ablation study. Finally we acknowledge the limitations of our research as a catalyst for future investigative directions. Our code is released \href{https://github.com/TangZJ/3iGS}{here}. \textbf{Acknowledgement} This study is supported under the RIE2020 Industry Alignment Fund - Industry Collaboration Projects (IAF-ICP) Funding initiative, as well as cash and in-kind collaboration from the industry partner(s). The computational work for this article was partially performed on resources of the \href{https://www.nscc.sg}{National Supercomputing Centre, Singapore}.


%
%
\bibliographystyle{splncs04}
\bibliography{main}
\end{document}